\documentclass[journal,comsoc]{IEEEtran}
\usepackage{mathrsfs}
\usepackage{graphicx}
\usepackage{multirow}
\usepackage[numbers,sort&compress]{natbib}
\usepackage[T1]{fontenc}
\usepackage{bbding}
\usepackage{color}

\ifCLASSINFOpdf
\else
\fi
\usepackage{amsmath}
\usepackage[cmintegrals]{newtxmath}
\begin{document}
%
\title{Towards Privacy-Preserving Person Re-identification via Person Identify Shift}
%
%
%

\author{Shuguang Dou, Xinyang Jiang, Qingsong Zhao, Dongsheng Li, Cairong Zhao
}

%
%

\markboth{> REPLACE THIS LINE WITH YOUR PAPER IDENTIFICATION NUMBER (DOUBLE-CLICK HERE TO EDIT) <}%
{Shell \MakeLowercase{\textit{et al.}}: Towards Privacy-Preserving Person Re-identification via Person Identify Shift}
%



\maketitle
\begin{abstract}
Recently privacy concerns of person re-identification (ReID) raise more and more attention and preserving the privacy of the pedestrian images used by ReID methods become essential. 
De-identification (DeID) methods alleviate the privacy issues by removing the identity-related mian of the ReID data. 
However, most of the existing DeID methods tend to remove all personal identity-related information and compromise the usability of de-identified data on the ReID task. 
In this paper, we aim to develop a technique that can achieve a good trade-off between privacy protection and data usability for person ReID.  
%
To achieve this, we propose a novel de-identification method specifically designed for person ReID, named Person Identify Shift (PIS). 
PIS removes the absolute identity in a pedestrian image while preserving the identity relationship between image pairs. 
By exploiting the interpolation property of variational auto-encoder, PIS shifts each pedestrian image from the current identity to another one with a new identity, resulting in images still preserving the relative identities.
Experimental results show that our method has a better trade-off between privacy-preserving and model performance compared to existing de-identification methods and can defend against human and model attacks for data privacy.
\end{abstract}


%
\IEEEpeerreviewmaketitle

\section{Introduction}
\IEEEPARstart{D}ue to the advances in cameras and web technology, it is easy to capture and share large amounts of video surveillance data, which facilitates the research and application of person re-identification (ReID) \cite{pr_book} in recent years. 
However, ReID has introduced severe concerns about personal privacy. 
For example, the data collected by ReID pipelines are usually video footage of surveillance cameras in public spaces, which could easily reveal highly confidential information such as daily individual whereabouts or personal activities. Aggregating unauthorized pedestrian data to a central server for the person ReID task may violate the General Data Protection Regulation (GDPR) \cite{GDPR}.



\begin{figure*}[t]
  \centering
  \includegraphics[width=\linewidth]{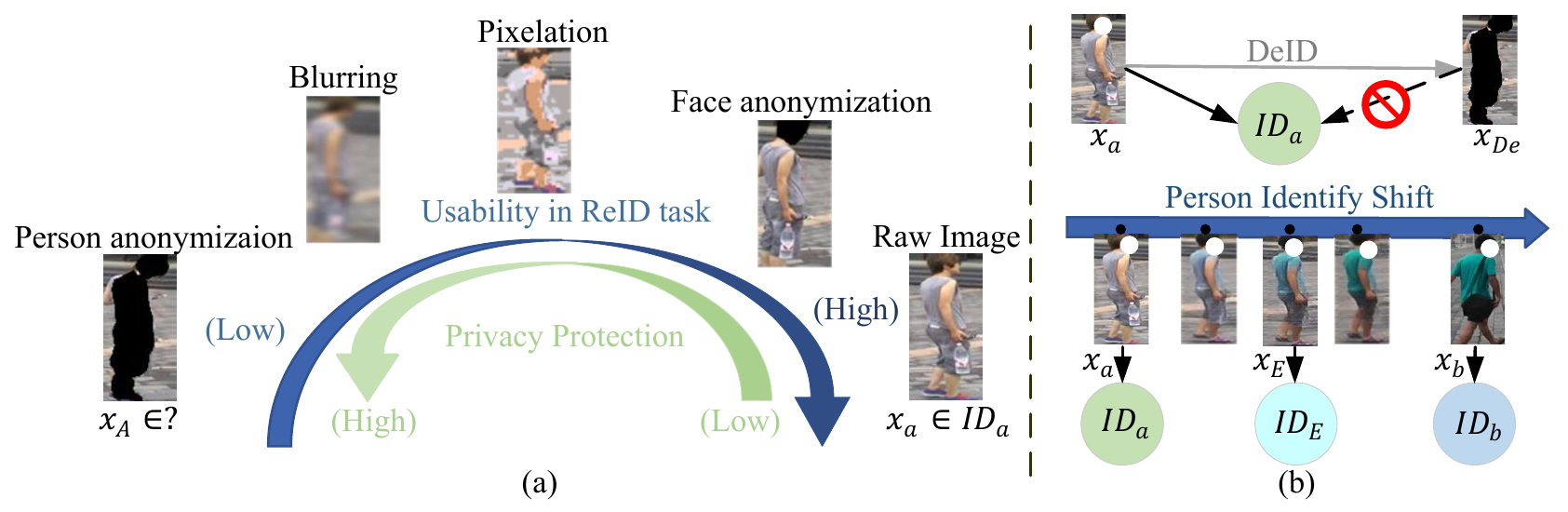}
  \caption{The challenge of privacy protection and our solution (a) The trade off between pravacy protection and Usability in ReID task . 
  (b) Person Identity Shift VS. Person De-identification}
  \label{fig:motivation}
\end{figure*}

In this paper, we focus on protecting privacy by removing the identity information from ReID images, namely person de-identification (DeID) \cite{pde-id,fde-id}. As shown in Fig. \ref{fig:motivation} (a), DeID aims to remove identifying information about the person with distortion methods like blurring, pixelation, crop, etc.
Most of the existing de-identification methods either achieve un-satisfactory privacy protection on ReID data or compromise the ReID performance. 
Firstly, many of the existing person DeID works remove privacy information from the face region by image distortion methods or generation models \cite{cvpr08_face_deid,jcst:ganforfacedeid,fde-id}.
However, the face is only one of the recognizable features of the human body; other biological features such as body structure, silhouette, gait, gender, and race could also reveal sensitive personal information. 
More importantly, although some DeID methods \cite{icip:hidingprivacyi,pde-id} can remove personal information of individuals other than face region features, these methods do not consider the usability of the de-identified data on the ReID task, leading to unsatisfactory ReID performance.  
It is extremely challenging to achieve a good trade-off between privacy protection and ReID data usability. 
This is because DeID methods aim to remove all personal identity-related information in each pedestrian image, but the ReID model relies on this removed identity information to conduct training and inference.
As a result, we seek to explore: Is it possible to protect privacy without compromising the usability of the data for ReID task? 

To resolve this challenging task, the proposed method is inspired by an important observation, which is person ReID models are learned to predict the similarity among pedestrian images, and hence the relationship between pedestrian image pairs (i.e., two images belong to the same identity or not) is much more important than the absolute identity of an image (i.e., who is the person in the image). 
We propose a method that only removes the absolute or real identity of the images, while still preserving the relationship between image pairs,  called person identify shift (PIS). PIS ensures images originally belonging to the same identity still have the same new identity after de-identification. 
Specifically, as shown in Fig. \ref{fig:motivation} (b), PIS transforms a set of images with the same identity to a new set of images with a different identity (i.e., identity shift). 
Inspired by the interpolation ability of variational auto-encoder \cite{ACAI}, this transformation is achieved by mixing the original image with several reference images in the latent representation space, leading to a generated image with a mixed identity. 
In addition, some existing public ReID datasets cannot be released due to privacy concerns. Our work provides a direction to publish the entire dataset after weak encryption by identity shifting, which can facilitate the flow of data and the development and community.


In conclusion, this paper takes one of the first steps toward addressing the challenge of privacy-preserving person re-identification via the following contributions:
\begin{itemize}
\item We propose a novel DeID method that achieves good trade-off between privacy protection and data usability for ReID, called Person Identity Shift (PIS), which shifts pedestrians' original identity to a new identity while preserving the relative identity relationship of image-pairs.    


\item We conduct detailed experiments to quantify the trade-off between the ReID performance and privacy protection ability to exist data de-identification methods and the proposed PIS. The results show that our method can outperform existing methods substantially on the commonly used ReID datasets. 

\end{itemize}


The rest of the paper is organized as follows. We introduce person ReID and general privacy protection methods, and privacy attack methods in Section \ref{sec:rela}. Section \ref{sec:method} presents the problem definition of privacy-preserving person re-identification and the details of Person Identify Shift. Experiments in Section \ref{sec:exp} demonstrate the effectiveness and safety of our approach. Finally, we conclude this paper in Section \ref{sec:con}.

\section{Related Work}
\label{sec:rela}
\subsection{Person Re-identification }
Person re-identification is a specific person retrieval problem that aims to seek the same person under non-overlapping cameras or at different times under the same cameras. Building a person ReID system need five main steps \cite{dl_pr}: (1) Collecting Raw data; (2) Generating Bounding Box; (3) Annotating Training Data; (4) Training Model and (5) Re-identification. In the above steps, there is some risk of privacy leakage. For example, in data collection, ReID data can be stolen by hackers when storing data centers. In training a ReID model with a distributed learning system, an attacker can also obtain private training data from a shared gradient \cite{Leakage}.

Existing person ReID methods mostly focus on performance improvements on public ReID datasets while neglecting privacy protection. DeID focuses only on privacy protection and ignores data availability for ReID task, i.e., they fail to maintain the ReID performance on the de-identified data. 
Different from the above tasks, the privacy-preserving person re-identification task proposed in this paper lies in balancing data availability and privacy protection.

\subsection{General Privacy Protection Methods}
We introduce two general privacy protection methods including confidential computing and differential privacy (DP) for deep learning.

\textbf{Confidential Computing. } To protect the privacy of data during the training process is confidential computing, such as Trusted Executive Environment (TEE), Homomorphic Encryption (HE) \cite{HE}, and Multi-party Secure Computation (MPC) \cite{MPC}. However, using them in modern deep learning environments is a challenge due to the high computational overhead and special settings such as public key infrastructure \cite{InstaHide}. 


\textbf{Differential Privacy. }
Training a model with ($\epsilon$, $\delta$)-DP \cite{our_data_ourselves} can protect the privacy of all training examples where ($\epsilon$, $\delta$)-DP means that the probabilities of outputting a model $M$ trained on two datasets $D$ and $D'$ that differ in a single example are close:
\begin{equation}
\label{eq:dp}
\operatorname{Pr}[M(D)] \leq \exp (\varepsilon) \cdot \operatorname{Pr}\left [M(D') \right]+\delta,
\end{equation}
where the ($\epsilon$, $\delta$) pair of Eq.\ref{eq:dp} quantifies the privacy properties of the DP-SGD, i.e., the smaller the $\epsilon$ , the better the privacy protection. 
However, it is challenging to apply differential privacy to deep learning. Abadi \textit{et al.} \cite{dp-sgd} propose DP-SGD to train deep learning models with DP. DP-SGD proved that privacy preservation for deep neural networks can be achieved with moderate cost in terms of training efficiency and model quality. However, DP-SGD only achieves good results on MNIST, dropping performance by about 20\% on Cifar10. 

\subsection{Privacy Attack Methods }
Since deep learning models always overfit the training data to some extent, the trained models can cause leakage of private training data \cite{PATE}. One common approach to attack model privacy are Model Inversion.
Model inversion infers training data from trained model or training process \cite{model_inversion,unintended_feature_leakage}. Hitaj \textit{et al.} \cite{GAN_leakage} train a GAN to generate prototypical samples of the training sets during the learning process. In Federated learning or Collaborative learning systems, people believe that sharing gradients will not expose private training data. However, Zhu \textit{et al.} \cite{Leakage} introduce the Deep Leakage from Gradient (DLG) to access private data in the training set from the publicly shared gradient. Zhao \textit{et al.} \cite{iDLG} found the ground-truth labels can be leaked from shared gradient and proposed the Improved DLG (iDLG) to extract accurate data from shared gradients.



\section{Method}
\label{sec:method}
\subsection{Problem Definition of Privacy Preserving Person ReID}
We define a ReID dataset as $\mathit{D_{reid}=\{x_i, y_i\}_{i\in N}}$ containing $N$ pedestrian images, where $x_i$ denote the $i-th$ pedestrian image and $y_i$ denotes its corresponding identity.
Privacy Preserving ReID (PPReID) aims at conducting Person ReID with minimum privacy information leakage. 
As a result, following the setting in \cite{PPAR}, to evaluate the trade-off between privacy protection and the ReID performance, PPReID involves two tasks, i.e., a target utility task $T_{u}$ that aims at achieving good ReID performance and a privacy budget $T_{p}$ task that prevents privacy leakage. 
In this paper, our novel de-identification method balances these two tasks by finding a data anonymization function $F_{anony}$ to transform raw data images into de-identified data $\hat x = F_{anony}(x)$. 

(1) \textbf{Target utility task}: For DeID methods, the target utility task  $T_{u}$ reflects how well the de-identified data can be used for ReID task. 
Intuitively, the performance of $T_{u}$ can be evaluated by the ReID model's performance trained on de-idenfied data, and a higher ReID performance indicates better usability of the de-identified data. 
Specifically, in our experiments, we choose several state-of-the-art ReID models to train on the de-identified data and evaluate their Rank-1 and MAP.

(2) \textbf{Privacy budget task}: The privacy budget task $T_{p}$ evaluates the ability of $F_{anony}$ to prevent data from leaking identity information.
$T_{p}$ in previous methods mainly focus on classification or detection tasks and can not be directly applied. 
As a result, we adapt them to ReID by reformulating the privacy attack as a retrieval task. 
Specifically, we assume that an attacker holds a query image of a target ID. The attacker determines whether this person exists in the de-identified dataset by using the query image to retrieve the correct target image from the de-identified ReID dataset. 
As a result, higher retrieval accuracy indicates a lower privacy protection effect, and the performance of $T_{p}$ is evaluated based on the retrieval performance metrics like Rank-1 or mean Average Precision (mAP).

$T_u$ and $T_p$ can be contradictory where higher $T_u$ performance may result in lower $T_p$ performance. Hence we propose a new general metric that considers the performances of both tasks and different importance of privacy and ReID, called PU-score (Privacy Utility Score):
\begin{equation}
\label{equ:balance}
    \text{PU-Score} = \frac{2}{\frac{I_u}{S_{T_{u}}} + \frac{I_p}{(1-S_{T_{p}})}},
\end{equation}
where $S_{T_{u}}$ is the average ReID performance, $S_{T_{p}}$ is the retrieval performance of the retrieval task in $T_{p}$, $I_u$ and $I_p$ denote the importance of the task in the PU-score and $I_u + I_p=2$.


\subsection{Overall Framework of Person Identify Shift}
\textbf{Overall Pipeline.}
Inspired by the interpolation ability of variational auto-encode (VAE) \cite{vae}, $F_{anony}$ is designed as a novel VAE that shifts the identity of the input images. 
Specifically, given a pedestrian image $x$, PIS de-identify it by mixing its latent representation with the images from other $k$ pedestrians, and decode the mixing embedding to a person with a new identity.  

\begin{figure*}[t]
  \centering
   \includegraphics[width=\linewidth]{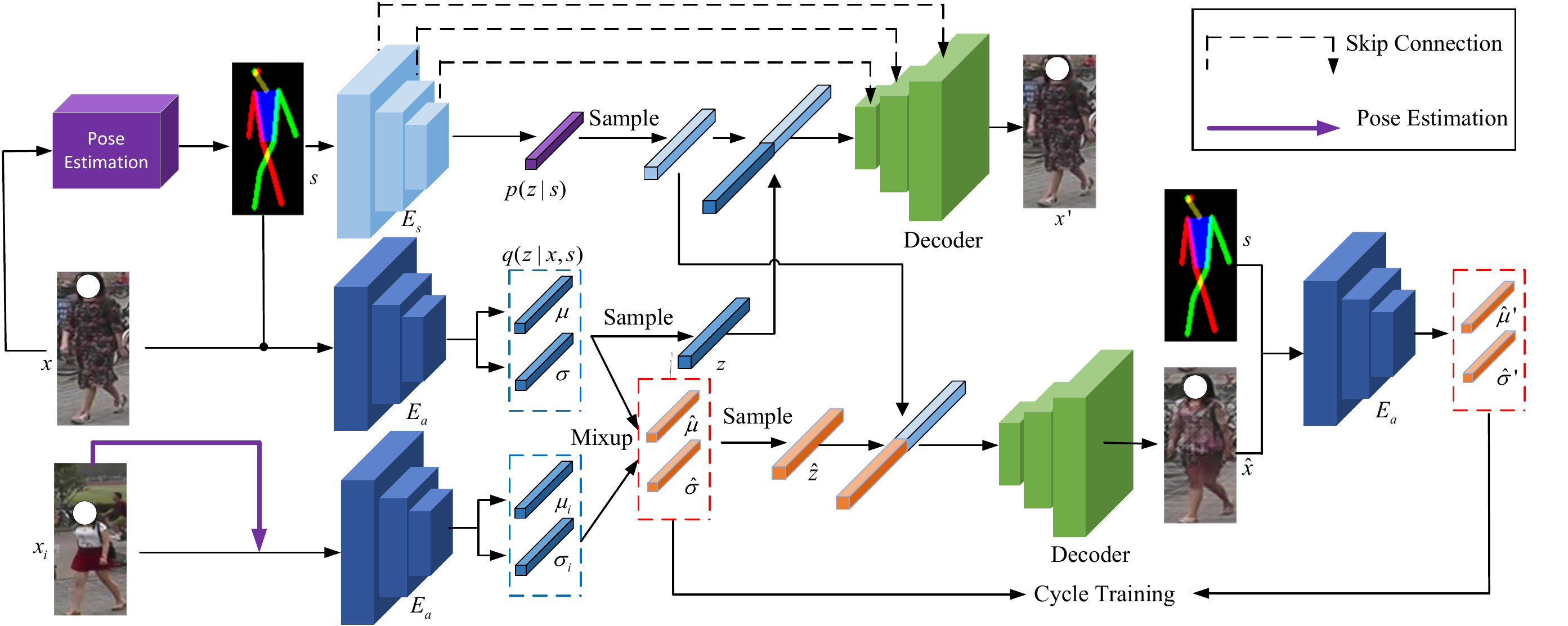}
   \caption{Framework of the proposed Person Identify Shift(k=2). }
   \label{fig:framework}
\end{figure*}

The overall pipeline of the Person Identify Shift is shown in Fig. \ref{fig:framework}. The PIS network takes the pedestrian image $x$ and its estimated pose $s$ as the input. The PIS network contains two Encoders and a Decoder. The body structure $s$ extracted from an image $x$ is fed into a prior-encoder $E_s$, which models the structure prior distribution $p(z|s)$, where $z$ is the latent variable.
Following \cite{VUNet}, conditioned on the image $x$ and body structure $s$, the appearance encoder $E_a$ models the posterior distribution $q(z|x,s)$. 
The Decoder models the likelihood $p(x|s,z)$, and reconstructs an new image $x'$ given the concatenation of latent variable $z$ and pedestrian pose $s$.

As shown in Fig. \ref{fig:framework}, the training pipeline of PIS mainly contains two branches. The first branch (upper row) conducts self-reconstruction, where image $x$ and its estimated pose $s$ is encoded and then reconstructed to the original image. 
The second branch (bottom row) generates new images based on a distribution mixture of $k$ images, where $k$ images are fed into the encoder, and their mixed representation $\hat z$ is sampled from a mixture distribution of their latent representation and fed into the decode to reconstruct an image with a new identity. Notably, PIS do not change the pose of the original image. This is because the DeID methods need to keep as much information as possible unchanged except for identity-related information, such as human behaviour that is important in the video.\looseness=-1


\textbf{Backbone Network.}
Similar to conditional U-Net \cite{UNet}, a skip connection is used between the structure decoder with the encoder. The Encoder and Decoder both use the same residual structure in ResNet \cite{ResNet} with $m$ residual blocks. Each residual block contains an ELU \cite{elu}  activation function and a $3\times3$ convolutional layer with a residual connection way. The residual blocks are connected with a downsampling layer, where the downsampling layer is a convolutional layer with a stride size of $2\times2$. After the downsampling layer, the feature map is reduced to 1/4 of the original. The decoder also uses $m$ residual blocks, while the sub-pixel convolution \cite{sub-pixel} between residual blocks is used as an up-sampling layer to change the feature map to four times the original. \looseness=-1

\textbf{Generating new ID by Mixture Distribution.}
Given a target image $x$ to be de-identified, its identity is shifted to a new one by mixing with $k - 1$ images $\{x_i\}_{i\in(2,..., k)}$ with different identities. 
Specifically, given target image $x$ and the other $k - 1$ images $\{x_i\}_{i\in(2,..., k)}$, the appearance representation of the generated new image are sample from a mixture of the $k$ distributions $q(z|x_i, s_i)$:
\begin{equation}
\begin{aligned}
    \hat z \sim \hat q(\hat z|\{x_i\}_{i\in(1,..., k)}, \{s_i\}_{i\in(1,..., k)}) 
    = \lambda_1 q(z|x, s) + \sum^{k}_{i=2}\lambda_i q(z|x_i, s_i),
\end{aligned}
\end{equation}
where $\hat q$ is a Gaussian mixture distribution containing $k$ sub-distributions, $\lambda_i$ is the probability that $z_a$ belongs to the $k$th sub-distributions, and $\hat z_a$ is sampled from $\hat q$. 
Then, the mixed latent representation $z_a$ are fed into the VAE decoder to generate an image $\hat{x}$ with a new identity, and hence $F_{anony}$ de-identifies $x$ by encrypting $x$ with $k - 1$ other images as follows:

\begin{equation}
\begin{aligned}
&F_{anony}(x) = D(\hat z, z\sim p(z|s)),
\end{aligned}
\end{equation}
where $D$ denotes the image decoder. 
Notably, the $k-1$ images used for creating a new ID can be seen as a ``random one-time private key" that weakly encysted the target image $x$.
Two constraints are made for the coefficients $\lambda_i$: (1) The sum of the coefficients $\lambda_i$ is 1, \textit{i.e.,} $||\lambda_i||_1=1$. (2) To prevent excessive coefficients from revealing private information of individual IDs, restrict all coefficients to be less than $d$.

\subsection{Optimization}
\textbf{Conditional VAE Loss.} The re-parameterization trick \cite{vae} is used to sample $z$ from the distribution, thus allowing backpropagation to optimize the parameters in $E_a$, $E_s$ and $D$. According to the standard Conditional VAE \cite{cvae}, the variational lower bound of
the model can be formula as:

\begin{equation}
\begin{aligned}
    \log p(x|s) \geq-KL((q(z|x, s) || p(z|s))
    +\mathbb{E}_{z\sim q_(z|x,s)}\left[\log p(x|s,z)\right],
\end{aligned}
\end{equation}
where $KL$ denotes the Kullback-Leibler divergence between the probability distributions $p$ and $q$ and $\mathbb{E}_{q_(z|x,s)}\left[\log p(x|s,z)\right]$ is the reconstruction term. The reconstruction term usually uses an L1-norm that often leads to ambiguous results. The perceptual loss, calculating the differences of activations on each layer of the pre-trained VGG model \cite{vgg}, can model perceptual images well. Thus, the reconstruction term in the conditional VAE loss is formulated as the perceptual loss:
\begin{equation}
\begin{aligned}
        \mathcal{L}_{cvae} = -KL((q(z|x, s) || p(z|s)) + 
        \alpha_{rec}\sum^m_i ||\mathcal{V}_i(x) -\mathcal{V}_i(x')||_1,
\end{aligned}
\end{equation}
where $\mathcal{V}_i$ is the $i$th layer of VGG-19, $\alpha_{rec}$ denotes a hyper-parameter that controls the weight of perceptual loss.

\textbf{Cycle Training for New ID Generation.} We denote the encrypted image generated from the embedding distribution mixture as $\hat x$. Since there is no corresponding ground truth image for $\hat x$, we propose to adopt a cyclical training loss to enforce better-mixed image generation results. Specifically, $\hat x$ is fed back into $E_a$ to get its embedding distribution $q(z|\hat x, s)$. Then, we minimize the Kullback-Leibler divergence between the embedding distribution of $\hat x$ and the original distribution mixture $\hat{q}$:
\begin{equation}
    \mathcal{L}_{cycle} = D_{KL}(\hat q||q(z|\hat x, s)).
\end{equation}


\textbf{Intra-Identity Appearance Consistency.} To make sure the images with the same identity have the same appearance, we use the image's corresponding identity label for supervised learning. The mean of appearance distribution $\mu_a$ is fed into an identity classifier containing a fully connected layer and a softmax layer.  We use the cross-entropy loss to supervise the prediction result.
\begin{equation}
    \mathcal{L}_{app} = SCE(fc(\mu_a), y),
\end{equation}
where $SCE$ denotes softmax cross-entropy loss and $fc$ denotes the fully connected layer.

In conclusion, the overall loss of PIS network is formulated as:
\begin{equation}
     \mathcal{L}_{total} = \mathcal{L}_{cvae} + \mathcal{L}_{cycle} +  \mathcal{L}_{app}.
\end{equation}
By jointly optimizing all the losses, both reconstruction quality and the relative identity of the encrypted images are ensured.




\section{Experiments}
\label{sec:exp}
First, we compare PIS with de-identification methods to verify that the proposed PIS is effective for PPReID. Second, we implement the user study to show the performance of our approach under human recognition attacks. Even though human recognition is far more expensive than model attacks, we also consider this possible attack. Third, if the attacker only has access to trained models or participates in distributed training such as federation learning, we mimic model reversal attacks to demonstrate the security of PIS. Finally, we compare with the differential privacy and provide visualization results.


\textbf{Dataset.} 
We conduct experiments on the two commonly used ReID datasets.
The details of the two ReID datasets are as follows:  1) Market-1501 \cite{market} contains 12,936 training images of 750 IDs, 3368 query images of 750 IDs, and 15,913 gallery images of 751 IDs. Following \cite{VUNet}, only 9939 training images of 730 IDs are used to train PIS. 
2) DukeMTMC-reID\cite{dukemtmc} contains 16,522 training images, 2,228 query images, and 17,661 gallery images. 

\textbf{Implementation Details.} The PIS framework is implemented by Tensorflow \cite{tensorflow}. All images of the training set are resized to $128\times 64$ and padding to $128\times 128$. The optimizer of PIS is Adam optimizer with a learning rate of 1.0e-3. The learning rate decays with training iteration. The batch size is 16 and the balance weight $\alpha_{rec}$ is 5. The multi-person pose estimation \cite{shape_est} is exploited to obtain the shape information.
In our experiments, we set k=2 and $\lambda_1=0.5$, where the mixed image is farthest from the original identities. Therefore, in our experiments, we set $\lambda_1$ to 0.5. In addition, to ensure the fairness of the ablation experiment, we used the same random selection strategy.


\begin{figure}[t]
  \centering
  \includegraphics[width=\linewidth]{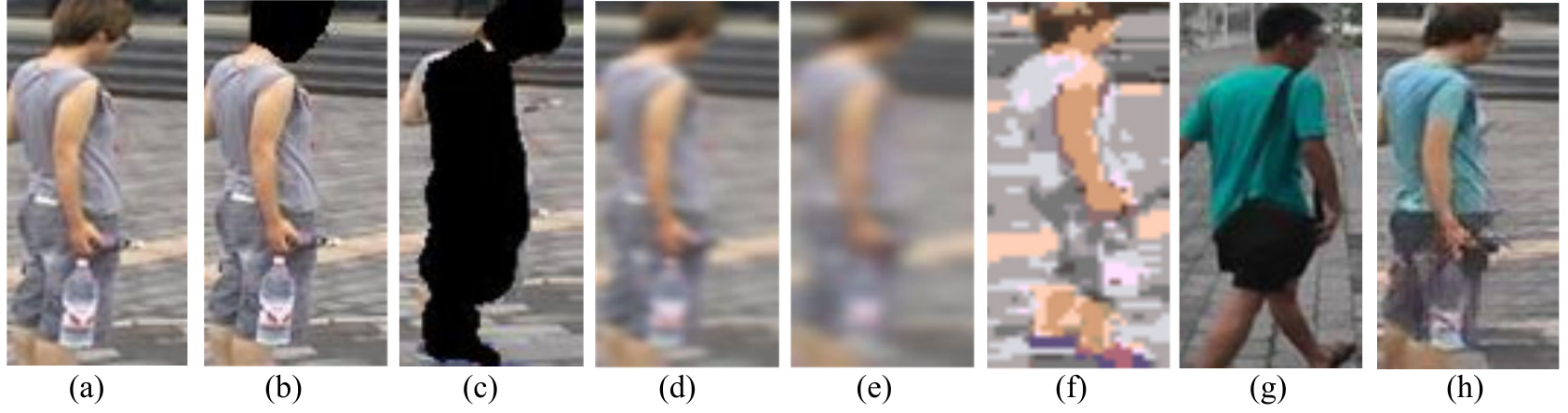}
  \caption{Comparing with different de-identification methods. (a) Origin image. (b) and (c) Remove pixels from head and body. (d) and (e) Gaussian blur with different size kernels. (f) Pixelation (g) Randomly selected images. (h) PIS (k=2)}
   \label{fig:de-id}
\end{figure}

\subsection{Trade-off between $T_{u}$ and $T_{p}$}
\label{sec:data}

As introduced in section 3.1, PPReID contains two tasks, namely the target utility task and privacy budget task. 
For target utility task $T_u$, three different state-of-the-art deep models (i.e., DenseNet121 \cite{densenet}, PCB \cite{partlevel-pcb} and ISP \cite{pixellevel-isp}) are trained on the de-identified data. 
The performance of  $T_{u}$ is the average MAP and Rank-1 of these three models. 
For privacy budget task $T_{p}$, the attack model-ISP is trained on a dataset separated from the training set of $T_{u}$ . 
The privacy attack is performed by using an attack model to query a set of target images from the de-identified training data.
The performance of $T_{p}$ is evaluated based on if the target images are successfully retrieved, and hence mAP and Rank-1 are used as evaluation metrics. 
Note that since PIS de-identify images by mixing identities, an attack is considered successful if either one of the mixed images is retrieved. 
The results of the trade-off between target utility and privacy budget are shown in Table \ref{tab:tradeoff} where $I_u = I_p = 1$.

\begin{table*}[t]
\caption{Performance comparison between PIS and existing image de-identification methods in terms of target utility (in \%) $T_u$, privacy budget (in \%) $T_p$ and PU-Score on Market-1501. Blur (S) and (L) denotes the use of $11\times5$ and $21\times11$ Gaussian kernels.}
\label{tab:tradeoff}
\centering
\begin{tabular}{l|llllll|llll|ll}
\hline
\multirow{2}{*}{Methods} &
  \multicolumn{2}{c}{DenseNet\cite{densenet}} &
  \multicolumn{2}{c}{PCB\cite{partlevel-pcb}} &
  \multicolumn{2}{c}{ISP\cite{partlevel-pcb}} &
  \multicolumn{2}{|c}{$T_{u} \uparrow$ } &
  \multicolumn{2}{c}{$T_{p} \downarrow$ } &
  \multicolumn{2}{|c}{PU-Score $\uparrow$ } \\
            & mAP  & R-1 & mAP  & R-1 & mAP  & R-1 & mAP      & R-1   & mAP  & R-1 & mAP  & R-1 \\ \hline
Origin        & 70.0   & 86.8 & 71.8 & 89.5 & 85.5 & 93.3 & \textbf{75.8} & \textbf{89.9} & 90 & 95.1 & 17.7 & 9.3  \\
Removing head & 69.5 & 86.7 & 68.6 & 87.1 & 79.4 & 90.7 & 72.5 & 88.2 & 87.8 & 94.0 & 20.9 & 11.2 \\
Removing body & 3.4  & 7.5  & 25.8 & 49.3 & 27.4 & 54.2 & 18.9 & 37.0 & \textbf{8.6}  & \textbf{21.2}  & 31.3 & 50.4 \\
Blur (S)      & 56.7 & 80.0   & 34.0   & 63.3 & 27.9 & 55.1 & 39.5 & 66.1 & 83.5 & 92.3 & 23.3 & 13.8 \\
Blur (L)      & 24.7 & 49.7 & 11.5 & 28.7 & 30.2 & 58.1 & 22.1 & 45.5 & 34.7 & 59.3 & 33.1 & 43.0 \\
Pixelation    & 43.7 & 68.2 & 48.7 & 73.3 & 65.4 & 83.7 & 52.6 & 75.1 & 56.2 & 76.9 & 47.8  & 35.3  \\ \hline
PIS (Ours)	& 46.4	& 70.0 & 	47.9 &	73.1&	61.3&	81.7&	51.9&	74.9& 35.1	& 48.5	&	\textbf{57.7} &	\textbf{61.0}
\\ \hline
\end{tabular}
\end{table*}

\textbf{PIS vs DeID.} We compare the PIS with the groups of person DeID approaches to demonstrate the advantages of PIS in balancing target utility and privacy budget. As shown in Fig. \ref{fig:de-id}, we compare with DeID-based methods including images with removing head and body, Gaussian blur with different sizes and pixelation. Firstly, removing all the pixels in the head region is not enough to defend against ReID model attacks, and intuitively the same conclusions go to other face de-identification methods. 

\begin{figure}[t]
  \centering
  \includegraphics[width=\linewidth]{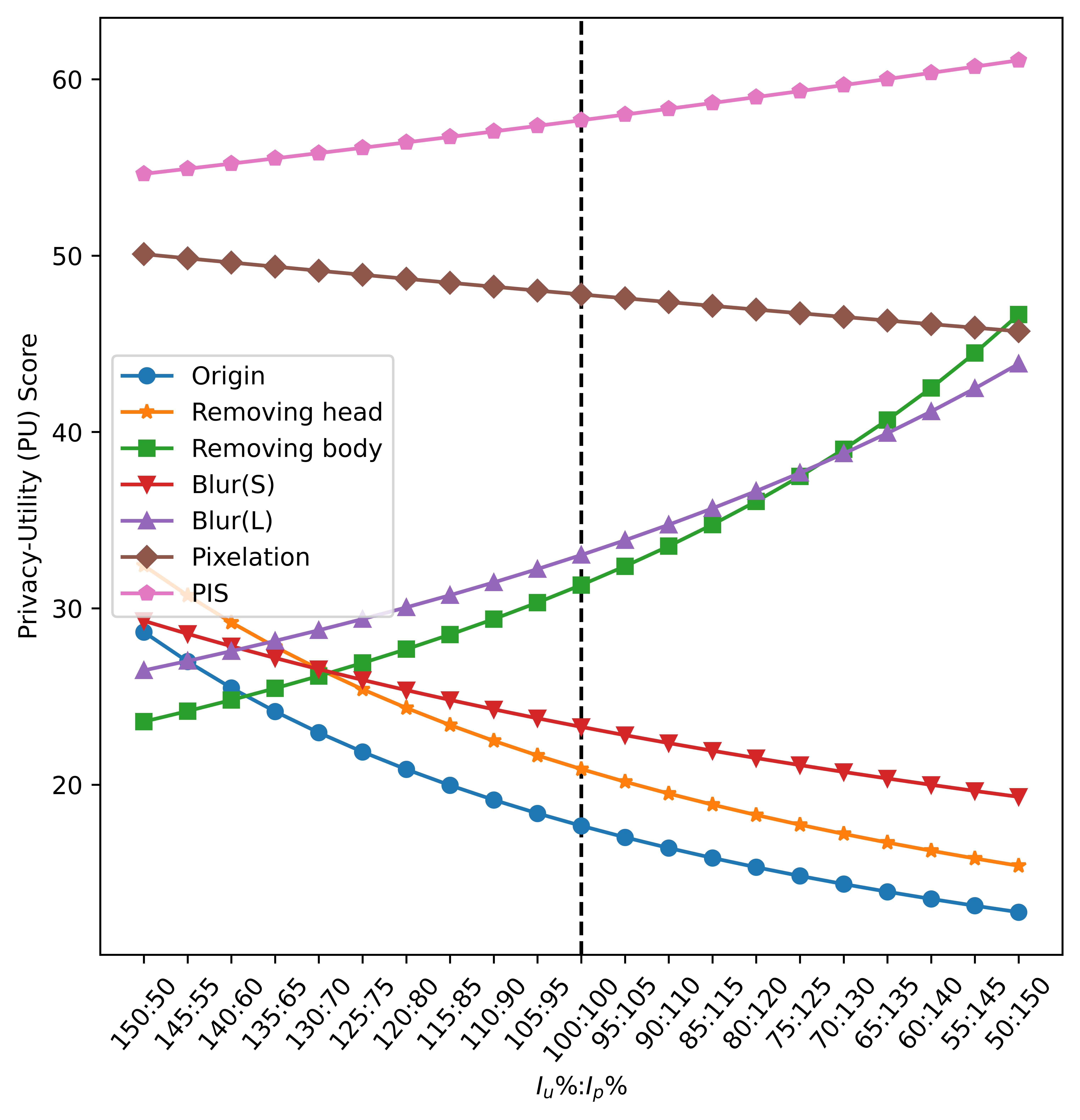}
  \caption{The trade-off between ReID performance and protecting data privacy under different $I_p:I_u$.}
  \label{fig:balance}
\end{figure}

Removing pixels from the human body from the video is the most effective way to protect individual privacy, i.e., mAP of 8.6\% under the ReID model attack. However, this method destroys the vast majority of information about the pedestrian in the image. Minimal blurring does not protect privacy, while severe blurring brings a loss of utility. This approach likewise fails to balance $T_{u}$ with $T_p$. Pixelation also does not perform well in the balance metric PU-Score. Compared with the above methods, the mAP and Rank-1 of PIS in PU-Score are much better than those DeID methods. 
Moreover, we also compare PIS and other methods under different privacy/ReID performance importance weight.  
As shown in Fig. \ref{fig:balance}, PIS is still far superior to other methods under all privacy-performance importance weight.

\begin{table*}[t]
\caption{Ablation study. AR denotes adversarial regular term of ACAI  \cite{ACAI}, CT denotes Cycle Training and AC denotes Appearance Consistency. $T_u$ only uses the performance of PCB.}
\centering
\label{tab:ablation}
\begin{tabular}{cccc|llll|ll}
\hline
\multirow{2}{*}{Baseline} &
  \multirow{2}{*}{AR} &
  \multirow{2}{*}{CT} &
  \multirow{2}{*}{AC} &
  \multicolumn{2}{|c}{$T_{u} \uparrow$ } &
  \multicolumn{2}{c}{$T_{p} \downarrow$ } &
  \multicolumn{2}{|c}{PU-Score $\uparrow$ } \\ 
  &   &   &   & mAP & R-1 & mAP  & R-1 & mAP      & R-1   \\\hline
\checkmark &   &   &   & 45.6  & 70.9     & 36.3 & 50.8     & 53.2 & 58.1 \\
\checkmark & \checkmark &   &   & 45.5  & 73.0     & 44.4  & 65.4   & 50.0 & 46.9 \\
\checkmark &   & \checkmark &   & \textbf{51.0}  & 75.1     & 42.6 & 58.7   & 54.0 & 53.3 \\
\checkmark &   &   & \checkmark & 50.0  & \textbf{75.7}     & 39.9 & 53.3  & 54.6 & 57.8 \\
\checkmark&   & \checkmark & \checkmark & 47.9  & 73.1     & \textbf{35.1} & \textbf{48.5}   & \textbf{55.1} & \textbf{60.4} \\ \hline
\end{tabular}
\end{table*}

\textbf{Ablation study.} 
This paper proposes two novel losses to generate images with mixed identities while maintaining appearance consistency within the same identity mixture, namely the cycle training loss (CT) and appearance consistency loss (AC).
We conduct an ablation study to analyze how different losses affect the ReID performance and privacy protection. 
Our baseline is as a conventional conditional VAE \cite{VUNet} with only VAE loss. We also compare with ACAI \cite{ACAI} that adds the proposed adversarial regular term to the baseline. The results are shown in Table \ref{tab:ablation}. First, the adversarial regular term can enhance the interpolation ability of the AE model, but it is not applicable for PPReID because it greatly increases the privacy budget. 
Cycle training and Appearance Consistency are both effective in the PU-Score and are used together to reduce the privacy budget while enhancing the target utility.


\textbf{Cross-dataset vs Inside-dataset.} To verify the generalizability of PIS, we shift the identities in Market-1501 to identities in the DukeMTMC-reID dataset, i.e, Cross-Dataset PIS. Previous experiments are all Inside-dataset PIS; in this part, we compare the Cross-dataset PIS with Inside-dataset PIS. As shown in Table \ref{tab:cross-dataset}, there is a decrease in target utility and an increase in privacy burden for Cross-dataset PIS compared to Inside-dataset PIS. The impact of cross-dataset is not significant for PIS, so in practice, PIS can be used between private ReID datasets with large-scale public ReID datasets to increase security.

\begin{table}[t]
\caption{Inside-dataset PIS vs Cross-dataset PIS on Market-1501. $T_u$ only uses the performance of PCB.}
\centering
\label{tab:cross-dataset}
\begin{tabular}{l|llll|ll}
\hline
\multirow{2}{*}{Datasets} & \multicolumn{2}{c}{$T_{u} \uparrow$} & \multicolumn{2}{c|}{$T_{p} \downarrow$} & \multicolumn{2}{l}{PU-Score $\uparrow$} \\
      & mAP & R-1 & mAP  & R-1 & mAP  & R-1 \\ \hline
Inside & 47.9  & 73.1     & 35.1 & 48.5   & 55.1 & 60.4   \\
Cross  & 45.5  & 71.3     & 39.4 & 55.5   & 52.0 & 54.8  \\\hline
\end{tabular}
\end{table}

\begin{figure}[t]
  \centering      
  \includegraphics[width=\linewidth]{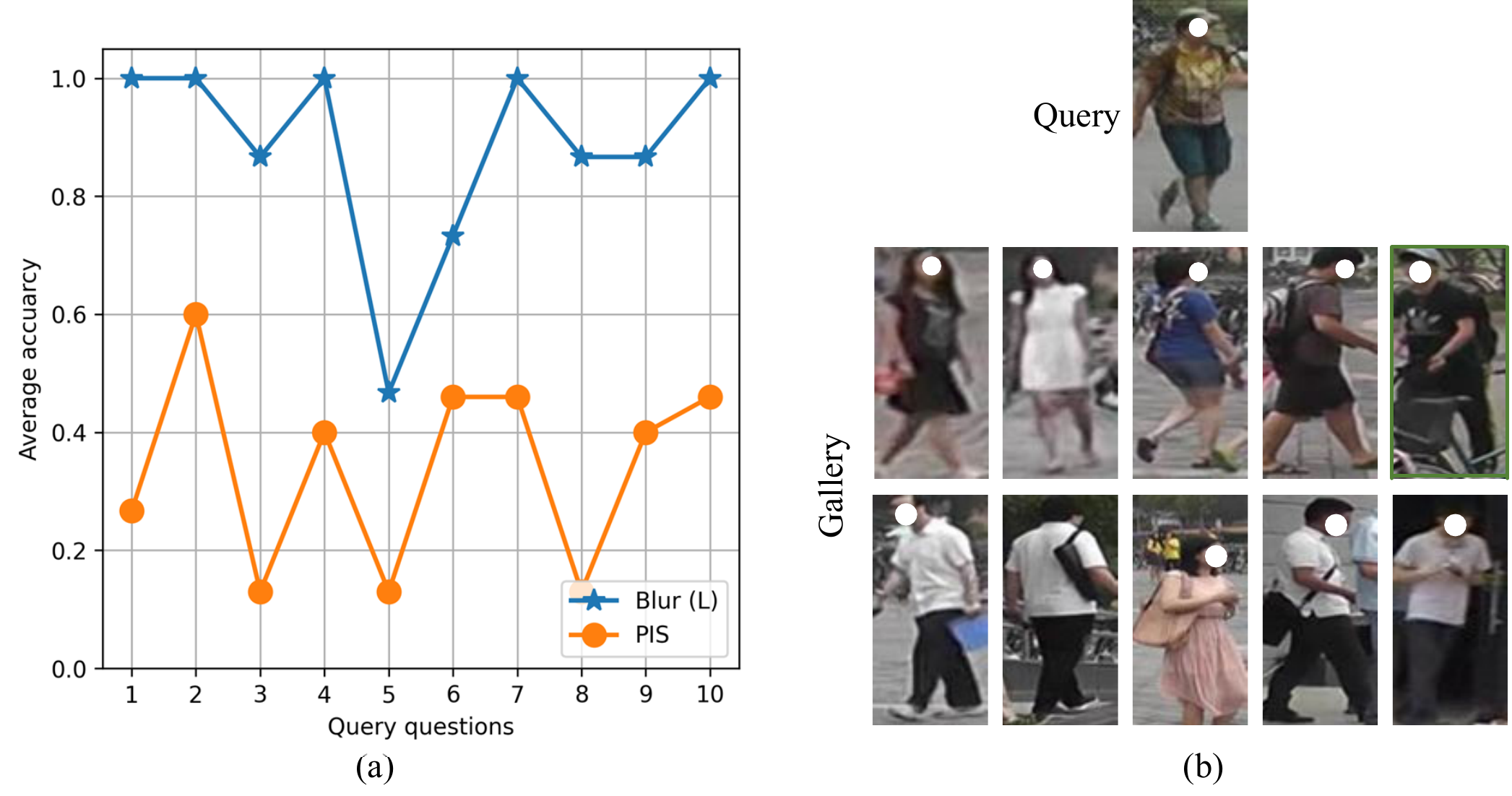}
  \caption{User study. (a) Average accuracy of 15 human observers on 10 query questions for PIS and Blur (L).  (b) Successful case of human attack for PIS. The pedestrian in the green box is the original identity corresponding to the Query.}
  \label{fig:user}
\end{figure}

\begin{figure*}[t]
  \centering      
  \includegraphics[width=\linewidth]{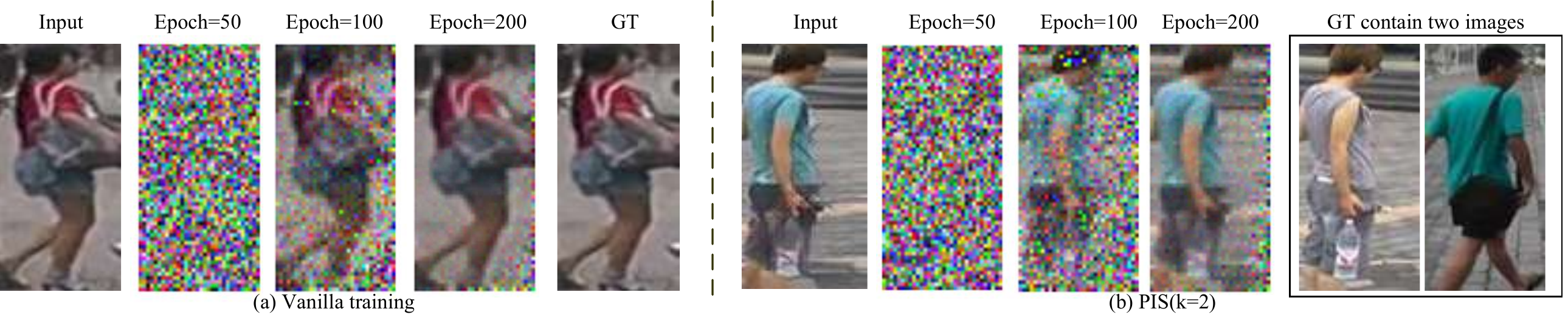}
  \caption{iDLG \cite{iDLG} steals training ReID images from shared gradient. (a) vanilla training. (b) Mixup training. (c) Training on images encrypted by PIS (k=2)}
  \label{fig:iDGL}
\end{figure*}
\subsection{User Study}
To further evaluate the privacy protection ability of PIS, we also conduct a user study to verify if the human can manually identify the image generated by PIS. Directly using human recognition ability to attack the de-identification method is one of the common methods, but human recognition ability is difficult to quantitatively state the efficacy of the method. Here we focus on the differences between human and model attacks. 

The average accuracy of 15 human observers for 10 query questions is shown in Fig. \ref{fig:user} (a). Multiple human observers attacked successfully on the same problem, and the successful case is as shown in Fig. \ref{fig:user} (b). There is only one person in the gallery with a hat, which is the reason for the success of the human attack. Our approach may fail to shift personal items such as backpacks and hats, based on which an attacker can obtain the original identity. Even so, PIS is far better than blur under human attack. And this problem of private items will be studied in our future work.


\subsection{Defend against Model Inversion Attack}
To simulate the possible existence of model inversion attacks, we use iDLG \cite{iDLG}, an improved version of DLG \cite{Leakage}, to steal private train set by the shared gradient.
According to the experimental setup of \cite{Leakage}, we made two changes to the ReID model. Since iDLG requires the model to be twice differentiable, we replace all the ReLU activation functions with Sigmoid and remove the strides as well. The L-BFGS\cite{L-BFGS} is used as an optimizer with a learning rate of 1.0 and 200 iterations. The attacked dataset is Market-1501\cite{market}.

The privacy leakage process is shown in Fig. \ref{fig:iDGL}. The stolen pedestrian images start as Gaussian noise, and the gap between the dummy data and the input to the model gets smaller and smaller as the training iterates. For all three training methods, iDLG can steal the input of the model from them. For the vanilla training method, iDLG directly steals the corresponding private data. Even with Mixup training, the stolen inputs can be used to isolate the mixed private data \cite{InstaHide}. Unlike the above methods, the PIS uses encrypted images to train the model, and the input to the model is not the real private data set, making it impossible for the attacker to directly steal the private data.

\label{sec:model}
\subsection{PIS vs Differential privacy}

PIS and Deferential Privacy is not directly comparable in terms of privacy protection, because one directly encrypts training data while the other focuses on preventing model leaking training data information. Following the setting of \cite{InstaHide}, we only report the target utility performance comparison between PIS and DP-SGD.

\textbf{Comparison with DP-SGD.} DP-SGD \cite{dp-sgd} implements differential privacy in deep learning by clipping the gradient first before adding noise during model training.  The BatchNorm \cite{bn} layer computes the mean and variance of the batch to create a dependency between samples in a batch that violates privacy. To implement DP-SGD in person ReID, we use Opacus \cite{opacus} to modify the ReID model by replacing all the BatchNorm (BN) layers in ReID models with GroupNorm (GN) \cite{gn}. As shown in Table \ref{table:PISvsDP}, our method shows good performance on ResNet50 and PCB without GN. In contrast, DP-SGD essentially fails to converge on the ReID dataset even with $\epsilon$ set to 50 (a very small noise) or changing learning rate. 
Moreover, another advantage of our method over the DP-based method is that no modifications to the regularization layer or the activation function \cite{tempered_sigmoid_for_dp} of the model is required.

\begin{table}[t]
\centering
\caption{Performance comparison between PIS and DP-SGD in terms of target utility on Market-1501. The GN indicates whether to replace BatchNorm layer with GroupNorm layer.}
\label{table:PISvsDP}
\begin{tabular}{llllll}
\hline
\multirow{2}{*}{Methods} & \multirow{2}{*}{GN} & \multicolumn{2}{c}{ResNet-50 \cite{ResNet}} & \multicolumn{2}{c}{PCB \cite{partlevel-pcb}} \\ 
    &   & mAP & Rank-1 & mAP  & Rank-1 \\ \hline
DP-SGD \cite{dp-sgd}  & \checkmark & 3.7 & 10.6   & 4.5  & 17.6   \\
PIS (Ours) & x &   46.5  &  67.0      & 47.9 & 73.1  \\ \hline
\end{tabular}
\end{table}

\begin{figure*}[t]
  \centering      
  \includegraphics[width=\linewidth]{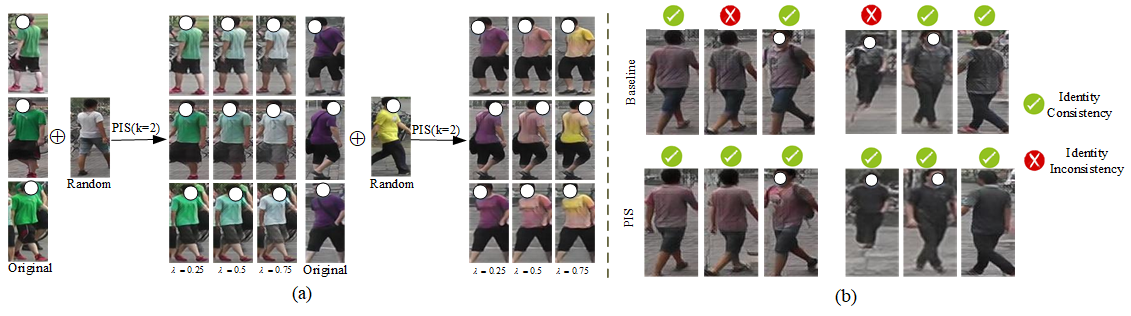}
  \caption{Examples of Person Identify Shift. (a) Visualization results of PIS at k=2 with different $\lambda$. (b) Intra-class consistency comparison within shift identities}
  \label{fig:vis}
\end{figure*}

\textbf{Comparison with adding random noise to appearance codes.} 
Another typical approach to achieve differential privacy is to add Laplace noise directly to the image. In this section, We add the Laplace noise to the appearance coding of pedestrian images and compare the structured appearance noise $n_A = \sum^{k}_{i=2}\lambda_i E_a(x_i)$ with the random Laplace noise $n_L$ where the mean of $n_A$ is equal to $n_L$. As shown in Table \ref{tab:laplace_noise}, adding random Laplace noise does not make the encrypted identity well away from the original one.

\begin{table}[t]
\caption{Performance comparison between Appearance noise and Laplace noise in terms of target utility (in \%) $T_u$, privacy budget (in \%) $T_p$ and PU-Score on Market-1501. $A$ denotes Appearance noise and $L$ denotes Laplace noise. $T_u$ only uses the performance of PCB.}
\centering
\label{tab:laplace_noise}
\begin{tabular}{c|llll|ll}
\hline
\multirow{2}{*}{Noise} & \multicolumn{2}{c}{$T_u \uparrow$} & \multicolumn{2}{c}{$T_p\downarrow$} & \multicolumn{2}{|c}{PU-Score$\uparrow$} \\
                & mAP  & R-1 & mAP  & R-1 & mAP      & R-1   \\ \hline
$A$ & 47.9 & 73.1   & 35.1 & 48.5   & 55.1 & 60.4 \\
$L$ & 52.4 & 77.0   & 60.0   & 78.7   & 45.4 & 33.4 \\ \hline
\end{tabular}
\end{table}

\subsection{Visualization Results}
The visualization results of PIS are shown in Fig. \ref{fig:vis}. Each row in Fig. \ref{fig:vis} (a) shows the results of applying PIS to mix an image of a target identity with a random reference image at the different mixing weights. 
We observe that when $\lambda$ is equal to 0.25 or 0.75, the encrypted image preserves a relatively large portion of the information from the mixing image with a higher mixing weight. When $\lambda$ is set 0.5, its appearance is dissimilar to either fixing images, and hence PIS can create a fictitious identity with a consistent style based on the existing training set. 
By comparing the generated images in each column of Fig. \ref{fig:vis} (a), we observe that the identity feature of the images in each column is very similar, and hence the relative identity is preserved for ReID training. 
We further compare PIS with the baseline method in Fig. \ref{fig:vis} (b), where the first three and last three images in each row use the same mixing configuration and generate images with the same identity. We observe that images generated by PIS are better at preserving intra-identity similarity than baseline.
PIS also achieves higher-order semantic interpolation, and the visualization result of the mutual shift between the four identities is shown in the Appendix.


\section{Conclusion}
\label{sec:con}
This paper takes the first step towards privacy-preserving person re-identification. 
We propose a novel privacy-preserving method called Person Identify Shift.
PIS generates weakly-encrypted and human-readable pedestrian images by shifting each image to a new image with a different identity. 
Since generated images with the same identity still have high appearance similarity, PIS can preserve the relative identity and hence is suitable for the ReID task. The experiments show PIS achieves a better trade-off between privacy protection and ReID performance.

\ifCLASSOPTIONcaptionsoff
  \newpage
\fi


\bibliographystyle{IEEEtran}
\bibliography{main}
\end{document}